\documentclass[a4paper, 10 pt, conference]{ieeeconf}

\IEEEoverridecommandlockouts                              

\usepackage{graphicx}
\usepackage{epstopdf}
\usepackage{multirow}
\usepackage{subcaption}
\usepackage{cite}

\usepackage{soul}
\usepackage{cancel} 

\usepackage{xcolor}
\usepackage{mathtools}
\usepackage{amsmath}

\newtheorem{problem}{Problem}
\newtheorem{remark}{Remark}
\newcommand{\e}[1]{\mbox{e}^{#1}}

\begin{document}
%
\title{\LARGE \bf A Data-Driven Slip Estimation Approach for Effective Braking Control under Varying Road Conditions}
%
%
%

\author{F.~Crocetti$^{1}$, 
             G.~Costante$^{1}$, 
             M.L.~Fravolini$^{1}$, 
             P.~Valigi~\IEEEmembership{Member,~IEEE}$^{1}$ 
\thanks{$^{1}$ Department of Engineering, University of Perugia, Via G. Duranti, 93, Perugia, Italy \tt\small\{francesco.crocetti, gabriele.costante, mario.fravolini, paolo.valigi\}@unipg.it}
\thanks{This work was funded by Clean Sky 2 Joint Undertaking (CS2JU) under the European Union's Horizon 2020 research and innovation programme, under project no. 821079, E-Brake}
}

\maketitle
\thispagestyle{empty}
\pagestyle{empty}

\begin{abstract}
The performances of braking control systems for robotic {platforms}, e.g., assisted  and autonomous vehicles, airplanes and drones, are deeply influenced by the road-tire friction experienced during the maneuver. Therefore, the availability of accurate estimation algorithms is of major importance in the development of advanced control schemes. The focus of this paper is on the estimation problem. In particular, a {novel} estimation algorithm is proposed, based on a multi-layer neural network. The training is based on a synthetic data set, derived from a widely used friction model.  The open loop performances of the proposed algorithm are evaluated  in a number of simulated scenarios. Moreover, different control schemes are used to test the closed loop scenario, where the estimated optimal slip is used as the set-point. The experimental results and the comparison with a model based baseline show that the proposed approach can provide an effective best slip estimation.

\end{abstract}

\section{Introduction}

The widespread diffusion of autonomous vehicles, drones, and in general, more advanced robotic terrain platforms, brings to extend research activities on traction and braking control systems too.
A  challenge that all these devices share is the estimation of the tire-road friction coefficient: the more the evaluation improves, the more the control system guarantees the behavior of the vehicle during the braking phase. Indeed, many effective slip control algorithms have been proposed, such as \cite{tanelli2008robust}, \cite{formentin2013data}, \cite{de2013wheel} and \cite{mirzaeinejad2018robust}. All of them require as input a slip set-point that can critically affects the performance of the entire system. 
Addressing the best (set-point) slip estimation problem usually involves the modeling of the tire dynamic and their relationship with the ground surface. 
In this context Artificial Neural Networks can improve the road estimation process: treated as smart sensors, their outputs can be used both as inputs for existing braking control algorithms and in combination with other estimation approaches.
Following this intuition, in this paper we propose a novel data-driven approach based on a Multilayer Neural Network ( Multi Layer Perceptron - MLP). 

In the literature, "slip-oriented" methods explore the effects of the vehicle dynamics steaming from a specific terrain surface: exploring the impact of friction on slip value (i.e., on the normalized difference between vehicle and wheel speeds)  to estimate the actual tire-road conditions. 

The typical friction model (see Section \ref{sec:problem}) assumes that the longitudinal tire-road force is given by $F_{x} = \mu(\lambda) F_{z}$, where $F_{z}$ is the normal force acting on the contact point between tire and road, and  $\mu(\lambda)$ is the normalized nonlinear friction curve, which is a lumped model for all the phenomena characterizing actual friction as a function of longitudinal slip.  
A common assumption of the estimation schemes based on nonlinear models, such as Pacejka \cite{bakker1987tyre} and Burckhardt \cite{burckhardt1993,kiencke2005automotive, khaleghian2017technical}, is that a single pair $(\lambda, \mu)$ is available at each time instant. 

\subsubsection{Model oriented}

In this framework,  least square and maximum likelihood approaches have been discussed in \cite{tanelli2009real,tanelli2008real}, with the purpose of estimating the whole tire-road friction curve. The idea is to estimate the parameters of the Burckhardt model, or to use a linearly parametrized approximation. 
A similar, linearly parametrized model is used  also in \cite{de2011optimal, de2012real} , where the basis functions are chosen according to a global optimisation problem, and a Recursive Least Square (RLS) approach is used to compute parameter estimates.
The proposal in \cite{tanelli2012combined} improves previous results by integration with a sliding mode observer to estimate vehicle acceleration.
The tire-road friction coefficient is also estimated by using an observer based approach \cite{baffet2007observer} and an \textit{Extended Kalman-Bucy Filter} \cite{ray1997nonlinear}.
Others approaches such as \cite{muller2003estimation}, \cite{aguado2018switched},\cite{bhandari2012surface}, \cite{rajamani2012algorithms},\cite{xia2016estimation} and \cite{de2013wheel} use a modified (i.e linear or linear approximation) version of the  $\mu-\lambda$ relation.

\subsubsection{Data-oriented}
The slip estimation problem has been addressed also by using machine learning tools and approaches.
A seminal work in this direction is \cite{pasterkamp1997application}, where a neural network has been trained by using data from the Pacejka model to estimate the friction coefficient $\mu$ and the slip angle $\theta$.  \cite{zhang2017hierarchical} uses a combination of a General Regression Neural Network (GRNN) and a Bayes filter to estimate the instantaneous friction coefficient $\mu$.
Other algorithms have also been explored for this problem, such as SVM  \cite{regolin2017svm} \space and deep learning \cite{song2018estimating}. Other problems and approaches are discussed in \cite{khaleghian2017technical,singh2018literature}, and \cite{guo2019review}. These data driven algorithms are typically combined with classical probabilistic filters such as EKF \cite{regolin2017svm}.


\subsubsection{Proposed approach}
The proposed approach relies on a novel data-driven strategy to model the relation between sequences of slip-friction pairs and optimal slip values. The solution is based on an MLP, trained on a synthetic dataset built from a set of Burckhardt curves.  The training phase has been designed to guarantee generalization, i.e, the capability of the networks to correctly react to input data not seen during training.  While the use of the Burckhardt model is widespread, to the best of authors knowledge, the approach used in this paper for network training is completely new, as well as the idea to estimate optimal slip instead of the actual surface model. Moreover, at runtime, the use of the proposed estimation scheme does not requires any knowledge of the friction model. The MLP takes as input data the vehicle and wheel speed, which are used to produce an estimate of a sequence of pairs ($\lambda$, $\mu$). The use of a sequence of pairs allows to embed the dynamic nature of the phenomena in the MLP input, and to smooth out high frequency variations. 
The generalisation capability acquired during the training phase allows the estimator to properly handle also situations with sudden surface variations, even in case of a sequence of those variations. Finally, differently from most of the literature, the estimated optimal slip is used as the set-point of a slip control scheme, achieving very satisfactory braking performances. Remarkable results were achieved also in case of road-surface variation during the maneuver. 

The paper is organized as follows. Section \ref{sec:problem} presents the dynamic model and formulate the problem. Section \ref{sec:estimation} propose and discuss the MLP architecture and the dataset used for the training phase. Section \ref{sec:slip_controller} briefly outlines the  control schemes, and the results are discussed in Section \ref{sec:exps}. Section \ref{sec:conclusions} draws conclusions and outline future work

\section{Problem definition}\label{sec:problem}

The dynamics of a vehicle, for the purpose of slip control, can be described by means of the Quarter-Car Model (QCM):
\begin{equation}
\begin{aligned}
J \dot{\omega} & =  r F_{x} - T_{w} \\
M \dot{v} & =  - F_{x}  
\end{aligned} \label{eq:QCM_model}
\end{equation}
where $\omega$ and $v$ are the wheel angular velocity and vehicle longitudinal speed, $J$ and $M$ are the associated momentum of inertia and mass,  $r$ is the wheel radius, and $T_{w}$ is the braking torque, i.e., the control signal.

The phenomena of interest in this paper depend on the longitudinal slip $\lambda$, which, during braking, is defined as:
\begin{equation}
\lambda := \dfrac{v - r \omega}{v} = 1 - r \dfrac{\omega}{v} . \label{eq:slip_definition}
\end{equation}

The key term in the QCM, for braking control, is the friction longitudinal force $F_{x}$, which describes the road-tire contact force.  
A widely adopted model for $F_{x}$  assumes dependence on the vertical force $F_{z}$ acting at the tire-road contact point, on the longitudinal slip $\lambda$ and the wheel side-slip angle $\theta$ according to the rule:
\begin{equation}
F_{x} = \mu(\lambda,\theta, \beta) \, F _{z} ,
\end{equation}
where the additional parameter vector $\beta$ characterizes the normalized friction function $\mu$ with respect to the specific type of road surface. In the following, it will be assumed the braking maneuvers will occur along a straight line. In these circumstances, the dependence of function $\mu$ on the wheel side-slip angle $\theta$ can be omitted.
A relevant situation where such an assumption is valid is the airplane landing phase. 

A largely used model for the normalised friction function $\mu(\lambda, \beta)$ is the static Burckhardt model \cite{burckhardt1993,kiencke2005automotive}, given by:
\begin{equation}
\mu(\lambda, \beta) = \beta_{1} \left(1- \e{\beta_{2}\lambda}\right) - \beta_{3}\lambda \, .
\label{eq:Burckardt_model} 
\end{equation}
\emph{The Reference Road Scenarios (see red curves in Figure \ref{fig:Sub_dataset_generated}) are: }\emph{Asphalt dry}   $(\beta_{1} = 1.2801, \, \beta_{2} = 23.99, \, \beta_{3} =0.52)$, \emph{Asphalt wet}   $(\beta_{1} = 0.857, \, \beta_{2} = 33.822, \, \beta_{3} =0.347)$, and \emph{Snow}   $(\beta_{1} = 0.1946, \, \beta_{2} = 94.129, \, \beta_{3} =0.0646)$.

Let denote with $\mu^{*}$ the \emph{optimal friction}, i.e., the maximum of the friction curve, and with $\lambda^{*}$ the \emph{optimal slip}, i.e., the associated slip value. The presence of such a local maximum implies that, for each road type, there is a single slip value yielding the best braking performance.

The slip function $\lambda$ can be assumed as the nonlinear output map of the dynamic model \eqref{eq:QCM_model}. Also, vehicle velocity $v$ changes in a slower manner than wheel speed $\omega$. Hence, by using $\lambda$ as a new state variable, in place of $\omega$, by assuming $v$ as a slow varying parameter, and by assuming  $F_{z} = M g$, the QCM dynamics can be rewritten as:
\begin{equation}
\begin{aligned}
\dot{\lambda} & = -\dfrac{1}{v} \left[ (1-\lambda) + \dfrac{M\,r^{2}}{J} \right] g \mu(\lambda) + \dfrac{r}{J \,v} T_{w} .
\end{aligned} \label{eq:slip_model}
\end{equation}

Based on the above considerations on the function $\mu(\lambda)$ and on the slip dynamics \eqref{eq:slip_model}, the problem of interest in this paper is the estimation of the optimal slip, and the use of such an estimate as the set-point of a slip controller. Hence, the objective of the paper is the solution to the following estimation problem.
\begin{problem}[Estimation of optimal slip]\em{
		Design a real-time  algorithm yielding the estimate $\hat{\lambda}^{*}$ of the optimal slip, using the measurements of the velocities $v$ and $\omega$, and the knowledge of the control input $T_{w}$.}
\end{problem}

\begin{remark}{\rm 
		The assumption of a known control signal is very reasonable whenever the control scheme is implemented in the same framework of the estimation scheme, where all the signals are available. In a number of cases, such as, for example, electrically powered braking systems, a load cell is available to measure braking torque. The scheme proposed in this paper can be easily extended to cover such a case.  
	}
\end{remark}

\section{Optimal Slip Estimation} \label{sec:estimation}


{A key challenge in most of the approaches to the estimation of the friction function $\mu(\lambda)$ relies on the non linear parametrisation  of the closed form models. This leads to a number of solutions, and notably to those based on approximate,  linearly parametrised models, such as, among others,\cite{tanelli2009real,tanelli2008real,rajamani2012algorithms,de2011optimal}.  }
\begin{figure*}[t]
	\begin{center}
		\includegraphics[width=0.9\linewidth, height=4.5cm]{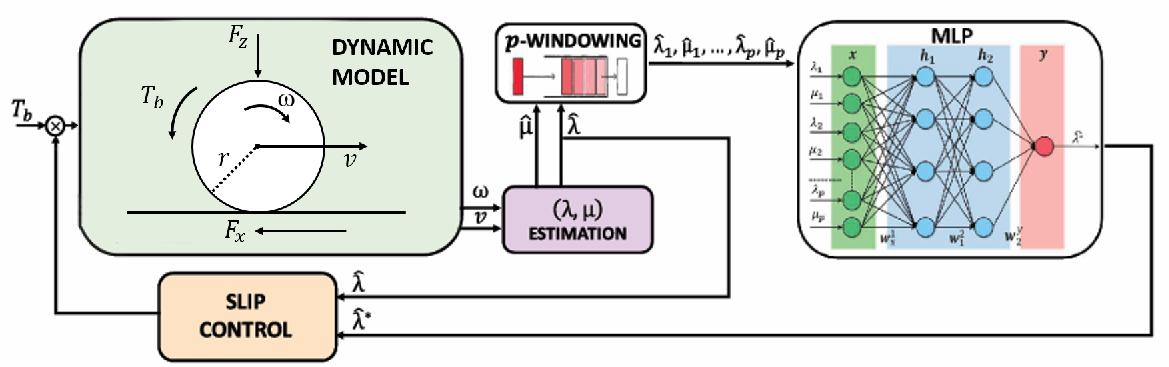} 
		\caption{The MLP estimation scheme, and the overall control scheme. \vspace{-10pt}}
		\label{fig:control_scheme}
	\end{center}
\end{figure*}


To tide over these issues, this work proposes a novel optimal slip estimation strategy, starting from a different hypothesis: the road-tire $\mu(\lambda)$ curve can be inferred by analyzing sequences of $(\lambda, \mu)$ pairs collected during the braking procedure. If this relation can be captured, it is reasonable to extend the estimation function to predict directly the optimal slip value $\lambda^{*}$, leaving to the model internal structures the task of representing the type of road surface. The use of a sequence of pairs, which appears as wholly new, allows to embed in the network the history of the braking, hence allowing to discriminate among the various possible friction curves. At the same time, this allow to partly filter out measurement and system noise. 

Driven by the previous considerations, this paper proposes a novel data-driven strategy to model the relation between sequences of slip-friction pairs and optimal slip values. The solution is based on a multilayer neural network (Multi Layer Perceptron-MLP), whose input features are vectors containing $(\lambda, \mu)$ pairs:
\begin{equation}
 \mathbf{x}=\{ \lambda_1, \mu_1, \lambda_2, \mu_2, \dots, \lambda_P, \mu_P  \} \label{eq:features} 
\end{equation}
with $P$ denoting the number of pairs, i.e., the window length.

The complete estimation algorithm is depicted, in the framework of the overall scheme, in Figure  \ref{fig:control_scheme}:
based on the measurements of longitudinal and angular velocities, and knowledge of the system parameters, an estimate of the corresponding $\lambda$ and $\mu$ values is derived by model inversion. This is a common approach in the QCM based literature\cite{de2013wheel, de2012real} while a more robust version could use a state observer, such as the one proposed in \cite{tanelli2012combined}. Once $P$ slip-friction pairs are acquired, they are concatenated and provided as input to the MLP, which outputs the estimated optimal slip. 
The MLP is trained in a supervised manner, by using ground truth optimal slip values $y=\lambda^{*}_{GT}$. 

\subsection{Dataset Construction} \label{sec:dataset_construction}

In order to train the MLP, it is necessary to collect a set of $N$ training samples $\mathcal{D}=\{(\mathbf{x}_1, y_1), (\mathbf{x}_2, y_2), \dots, (\mathbf{x}_N, y_N)\}$, where $\mathbf{x}_i$ are sequences of slip-friction pairs in \eqref{eq:features}, {and $y_{i}$ are the associated optimal slip values.}
This dataset could be obtained by sampling the reference curves \textit{i.e.,} Asphalt dry, Asphalt wet, and Snow. However, such a data set would be not sufficient to allow the MLP to generalize with respect to other possible scenarios. Hence, different values for the parameters $\beta_1$, $\beta_2$ and $\beta_3$ are explored, to generate different slip-friction curves. In particular, for each parameter $\beta_j$, an interval $B_j$ of possible values has been defined, in order to cover all the reference surfaces.

The space $(B_{1} \times B_{2} \times B_{3}) \in \rm I\!R^3$, defined as the friction cube, is sampled by the following two strategies in order to generate different road scenarios. First, $N_{diag}$ road surfaces are sampled on the cube diagonal. This choice stems from the observation that the reference surfaces lie close to the diagonal. This makes it possible to simulate road scenarios that resemble the reference ones. Secondly, to represent more different road conditions, $N_{hyp}$ curves are sampled in the entire cube by using the Latin Hypercube Algorithm \cite{helton2003latin}.

\begin{figure}[h!]
	\begin{center}
		\includegraphics[width=0.97\columnwidth, height=5.2 cm]{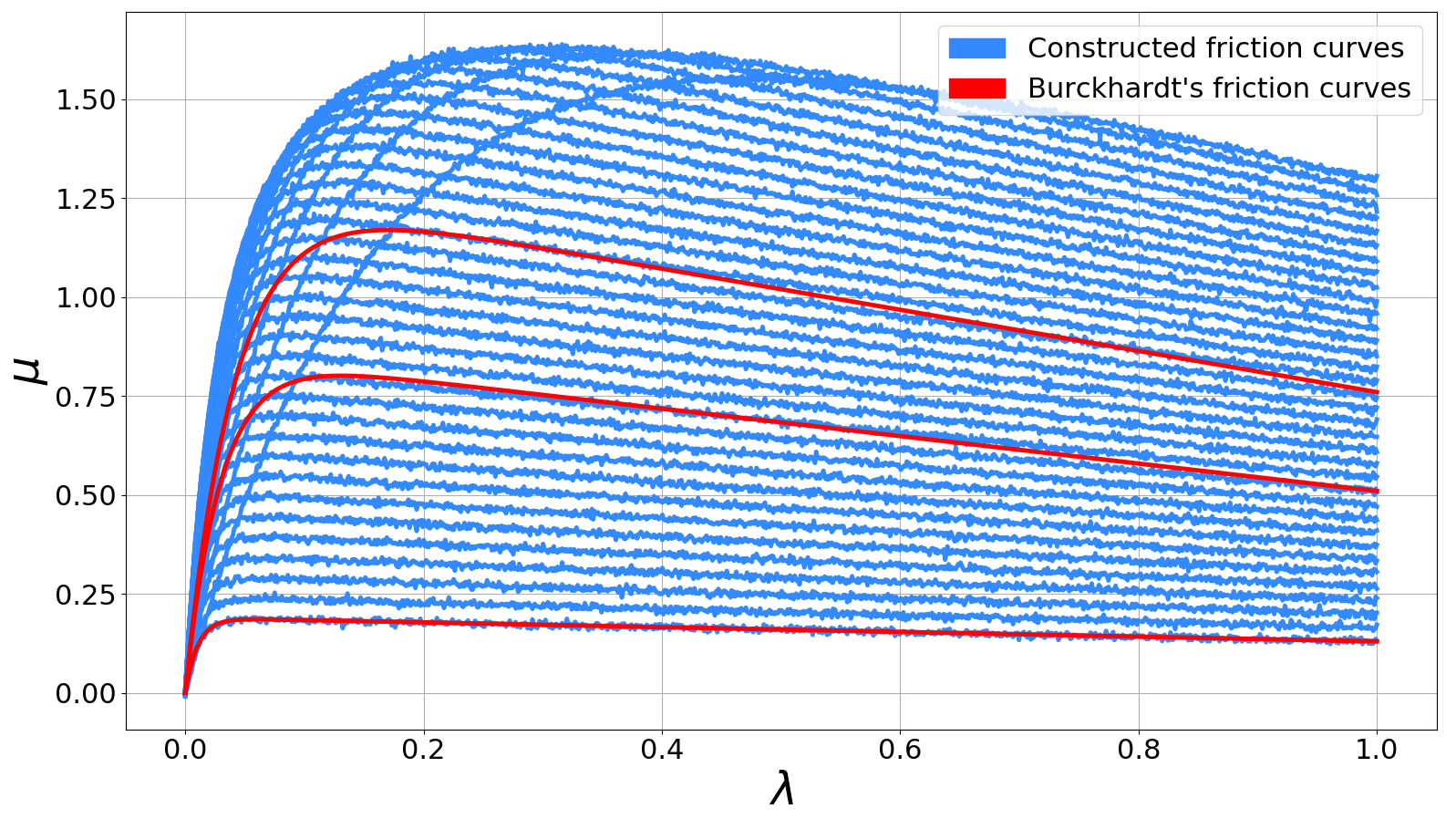} 
		\caption{Complete set of curves. \vspace{-0.7cm}}  \label{fig:Sub_dataset_generated}
	\end{center}
\end{figure}

To generate the inputs to the MLP, each slip-friction curve is discretized with $1000$ points along the slip range $\lambda \in [0,1]$. Afterwards, a sliding window of fixed size is used to select $P$ pairs $(\lambda_i, \mu_i)$ ($P=50$ in this study, based on extensive experimental tests) and build the feature vector. Each window is then associated to the optimal slip value $\lambda^{*}_{gt}$, which is computed on the basis of the closed form model for $\mu$.
The measurement noise is modeled by an AWG noise $\mathcal{N}(0,\,\sigma^{2})$ acting on the $\mu$ values, with $\sigma=0.005$.  The obtained set of curves are depicted in blue in Figure \ref{fig:Sub_dataset_generated}. It is stressed that these data are only used on the training stage, and not during the test phase. 

\subsection{Network Structure} \label{sec:network_structure}

The neural network architecture used to achieve optimal slip estimation is designed with Rectified Linear Unit (ReLU) non linearities as hidden layer activations and is trained by optimizing the \textit{Mean Squared Error (MSE)} loss function \cite{bishop2006pattern}. The hyper-parameters (\textit{i.e.,} number of hidden layers, number of neurons, learning rate and optimizer) are selected by using standard cross-validation procedures. The best performances are achieved 
 with a stochastic gradient descent optimizer \cite{bishop2006pattern} with a learning rate of $0.01$, and two hidden layers with $250$ neurons each.

\section{Slip Controller} \label{sec:slip_controller}

In order to evaluate the slip estimation scheme, a closed loop slip control system has been considered. Two slightly different controllers have been evaluated, a PI regulator, and a Sliding Mode Controller (SMC), based on the approach proposed in \cite{de2013wheel}. Both controllers assume that an external braking signal $T_{b}$ is generated by the vehicle pilot. The controller purpose is to reduce the braking effort in order to keep the slip value to the given reference value. Hence, the overall braking signal is given by $T_{w}  = T_{b} u$, where $u$ is the closed loop control signal. The controller set-point is given by the estimate of the optimal slip provided by the MLP.
The sliding mode controller, discussed in  \cite{de2013wheel},  comprises a switching term to robustly drive the system to the sliding surface, and an integral term to guarantee zero steady state error. 
The overall control scheme is given by:
\begin{equation}
\begin{aligned}
T_{w} & = T_{b} u , \qquad u    = \dfrac{1}{2} - \beta \mbox{\rm{sat}}\left(\dfrac{e + k_{0}\sigma}{\epsilon}\right) \\
s_{c} & =  e + k_{0}\sigma , \qquad 
e(t)  =  \hat{\lambda}^{*} - \lambda(t) \\
 \dot{\sigma} & = - k_{0} \sigma + \epsilon \mbox{\rm{sat}}\left(\dfrac{s_{c}}{\epsilon}\right)  
\end{aligned} \label{eq:SMC_CI_regulator}
\end{equation}

\section{Simulation results} \label{sec:exps}

\begin{figure*}[t]
	\centering
	\begin{subfigure}{0.3\linewidth}
	\includegraphics[width=\textwidth]{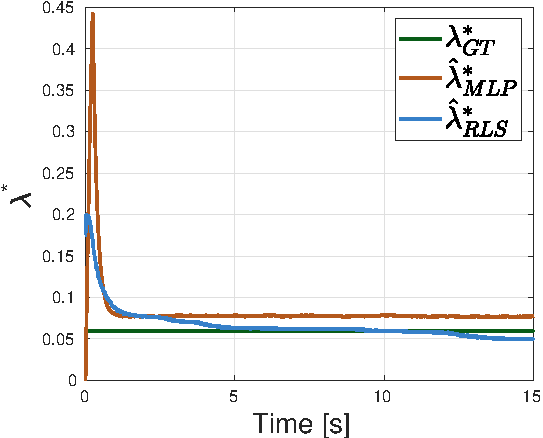}
	\caption{SNOW}
	\label{fig:open_loop_snow}
	\end{subfigure}
	\begin{subfigure}{0.3\linewidth}
		\includegraphics[width=\textwidth]{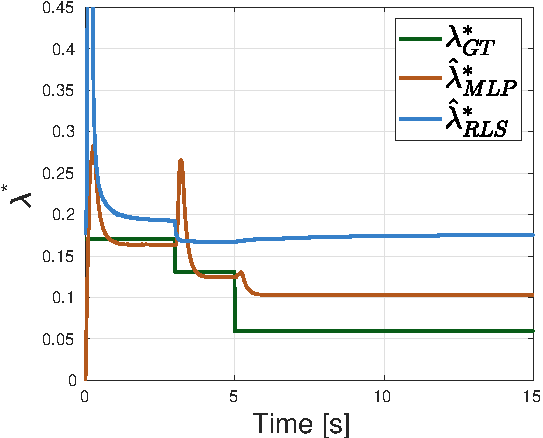}
		\caption{D$\rightarrow$W$\rightarrow$S}
		\label{fig:open_loop_DWS}
	\end{subfigure}
	\begin{subfigure}{0.3\linewidth}
		\includegraphics[width=\textwidth]{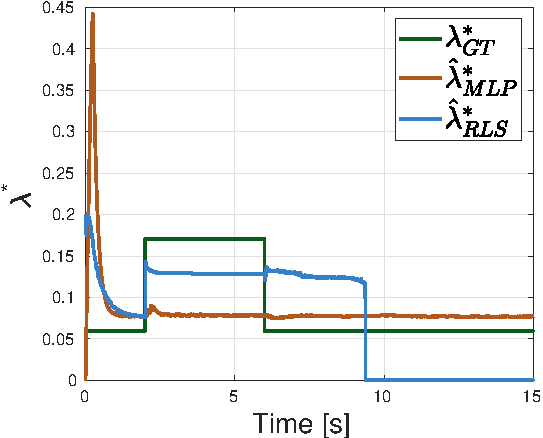}
		\caption{S$\rightarrow$D$\rightarrow$S}
		\label{fig:open_loop_SDS}
	\end{subfigure}

	\caption{Open loop braking braking maneuvers: time behaviour of true optimal slip, RLS and MLP estimate. }
	\label{fig:open_loop}
\end{figure*}

\begin{figure*}[t]
	\centering
	\begin{subfigure}{0.3\linewidth}
		\includegraphics[width=\textwidth]{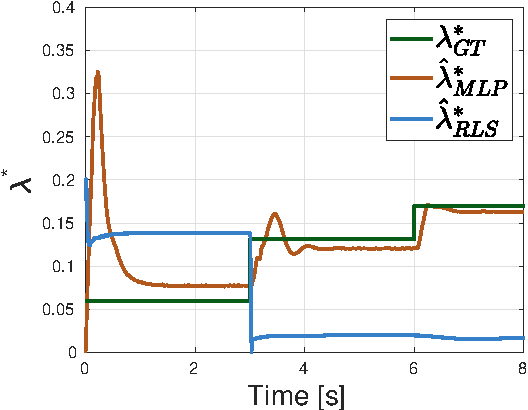}
		\caption{SMC S$\rightarrow$W$\rightarrow$D}
	\end{subfigure}
	\begin{subfigure}{0.3\linewidth}
		\includegraphics[width=\textwidth]{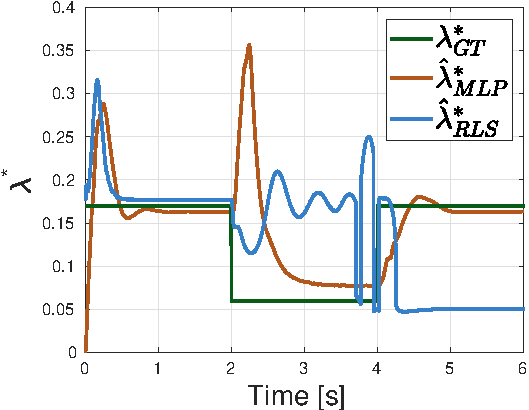}
		\caption{SMC D$\rightarrow$S$\rightarrow$D}
	\end{subfigure}
	\begin{subfigure}{0.3\linewidth}
		\includegraphics[width=\textwidth]{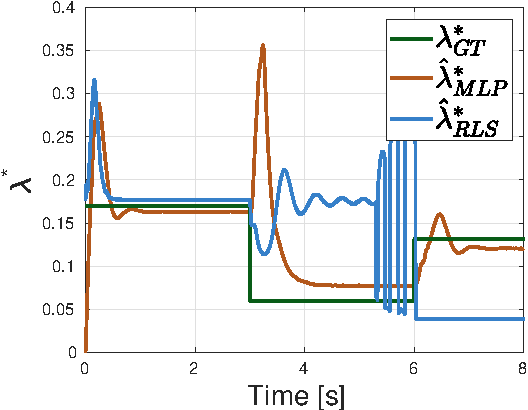}
		\caption{SMC D$\rightarrow$S$\rightarrow$W}
	\end{subfigure}
	
	\caption{Time behavior of braking maneuvers under closed loop control schemes. }
	\label{fig:closed_loop}
\end{figure*}

\subsection{Experimental settings}

To asses the benefits of the proposed approach, three different set of tests are provided.

\subsubsection{Optimal Slip Estimation Test}

First, the capability of the MLP model to predict the optimal slip value for a given road surface is evaluated. To fulfill this purpose, the set of curves generated as described in Section \ref{sec:dataset_construction} are divided into training, validation and test subsets.
It should be noticed that this choice allows to test whether the model is able to generalize with respect to unseen road surfaces or not. Furthermore, the three reference roads  are used as an additional test bench (since they are excluded from the training set). These curves are represented in Figure \ref{fig:Sub_dataset_generated}, Burckhardt's in red, constructed in blue. Quantitative evaluations are provided by computing the Root MSE (RMSE) \cite{bishop2006pattern} over the training and the test sets, and for the reference roads.
 
\subsubsection{Open Loop Test}

This case is evaluated by simulating the landing of an aircraft  over an unknown surface whose conditions change during the braking operation. 
In particular, a step signal simulating the pilot brake request $T_{b}$, is provided to the QCM, which returns wheel and vehicle speed (with $M= 1600\,(\text{Kg})$, $J=0.4500\,(\text{Kg}\cdot{\text{m}^2})$, and $r=0.3\,(\text{m})$). In these tests, the system operates in \textit{open loop} configuration, \textit{i.e., } pilot braking signal $T_{b}$ is directly applied to the vehicle, without the intervention of any slip control scheme.
In all the experiments, the initial aircraft speed is set to $80\,(\text{m/s})$, while the initial wheel velocity is set to simulate a null initial value for the slip, i.e., the case where braking starts after ground contact.
During braking, road conditions are changed with step-wise transitions.  The requested braking force $T_{b}$ is set to the value that experimentally gives  the best performance  for the initial surface, and it is not modified during manoeuvre. The tests are made by exploring transition between the three reference surfaces, \textit{i.e.,} Asphalt Dry (D), Asphalt Wet (W) and Snow (S).  It is important to stress that these  surface {transitions} are not used in the training phase of the MLP.
The performance of the proposed MLP predictor are compared against the approaches presented in \cite{de2011optimal,de2012real}, which rely on the RLS strategy. The RMSE between the ground truth and the estimated optimal slip during the entire braking operation is computed to provide a quantitative evaluation.

\subsubsection{Closed Loop Test}

The availability of optimal slip estimates allows to benefit from slip control schemes, aimed at regulating slip to such an optimal value
 (see Figure \ref{fig:control_scheme}).
The control schemes described in Section \ref{sec:slip_controller} have been used. It is stressed again that the focus of the paper is on the MLP estimator, hence control schemes are only used for the purpose of such a study.
Similarly to the case of the open loop test, the road conditions are changed during the experiment and the performance are evaluated measuring the RMSE between the ground truth and the estimated optimal slip, the required braking time and the traveled distance.

\subsection{Results and Discussion}

\subsubsection{Optimal Slip Estimation Results}
A first consideration about the generalization capabilities of the neural network can be made analyzing the RMSE scores achieved by the MLP listed in table \ref{tab:mlp_scores}.
\begin{table}[h]
	\centering
	\resizebox{0.93\columnwidth}{!}{%
		\begin{tabular}{c|c|c|c|c|}
			\cline{2-5}
			& Training & Validation & Test & Reference \\ \hline
			\multicolumn{1}{|c|}{$RMSE$} & 0.0463 & 0.0464 & 0.0290 & 0.0361 \\ \hline
		\end{tabular}%
	}

	\caption{MLP scores on the datasets\vspace{-10pt}}
\label{tab:mlp_scores}
\end{table}

The results obtained on the test data and the reference roads are comparable (in this case better) to the training set ones. By observing that the test and the reference roads are completely excluded during the training procedure, it is possible to conclude that the MLP is able to generalize with respect to "unseen" surfaces.
Moreover, by considering that a typical value for the optimal slip is $\lambda^{*}=0.15$, the error percentage of the MLP (with respect to the RMSE) is $19\%$ on average. As shown by the other tests, this error is reasonable when  the estimate is used as the set-point values of a slip control scheme.

\subsubsection{Open Loop Results}

The performance of the MLP estimator are compared against the RLS strategy proposed in \cite{de2011optimal,de2012real}. 
\begin{table}[htb]
	\centering
	\resizebox{0.5\columnwidth}{!}{%
		\begin{tabular}{|l|c|c|}
			\hline
			\multirow{2}{*}{Asphalt} & \multicolumn{2}{c|}{$RMSE$} \\ \cline{2-3} 
			& MLP & RLS \\ \hline
			D & \textbf{0.0071} & 0.0213 \\ \hline
			W& \textbf{0.0034} & 0.0216 \\ \hline
			S & 0.0175 & \textbf{0.0101} \\ \hline
			D$\rightarrow$S$\rightarrow$D & \textbf{0.0244} & 0.0415 \\ \hline
			S$\rightarrow$D$\rightarrow$S & \textbf{0.0314} & 0.0582 \\ \hline
			W$\rightarrow$D$\rightarrow$ W & \textbf{0.0510} & 0.0700 \\ \hline
			D$\rightarrow$W$\rightarrow$ D & \textbf{0.0131} & 0.0260\\ \hline
			W$\rightarrow$S$\rightarrow$ W & \textbf{0.0238} & 0.0312 \\ \hline
			S$\rightarrow$W$\rightarrow$S & \textbf{0.0226} & 0.0668 \\ \hline
			S$\rightarrow$W$\rightarrow$D & \textbf{0.0872} & 0.1294 \\ \hline
			W$\rightarrow$D$\rightarrow$ S & 0.0320 & \textbf{0.0227} \\ \hline
			D$\rightarrow$S$\rightarrow$W & \textbf{0.0206} & 0.0608 \\ \hline
			S $\rightarrow$D$\rightarrow$ W & \textbf{0.0536} & 0.1212 \\ \hline
			W$\rightarrow$S$\rightarrow$D& \textbf{0.0378} & 0.0473 \\ \hline
			D$\rightarrow$W$\rightarrow$S & \textbf{0.0401} & 0.1071 \\ \hline
		\end{tabular}%
	}
	\caption{Scores of RLS and MLP: open loop data}
	\label{tab:Openloop}
\end{table}
Tests are made both with fixed and  changing surface conditions during the aircraft braking (see Table \ref{tab:Openloop}). 

The MLP achieves lower estimation errors in most of the cases, with only a few exceptions where the RLS scores are better than the neural network, and all of them include the snow surfaces (see Figure \ref{fig:open_loop_snow}).
The motivation behind this slight MLP performance drop can be explained by observing that the optimal slip value for snow surface is close to zero. As shown in Figure \ref{fig:Sub_dataset_generated}, for small $\lambda$ values, all the curves exhibit very similar behaviours and, thus, it is more challenging for the MLP to handle ambiguities in these scenarios. This effect is further emphasized by the measurement noise, which makes the curves nearly indistinguishable for small slip values.
Nevertheless, even in those situations, the MLP performance are close to the RLS ones.
Figure \ref{fig:open_loop} provides a qualitative analysis of the open loop test with various transitions, where  $\lambda^*_{GT}$ denote the optimal slip values, while $\hat{\lambda}^*_{MLP}$ and $\hat{\lambda}^*_{RLS}$ the estimated ones. The MLP-based best slip estimator shows better performance with respect to the RLS, especially when road transitions occur. The spikes that affect the MLP estimation behavior during the road surface switch are caused by the transitory presence of inconsistent $(\lambda, \mu)$ pairs in the input vector to the neural network. In particular, the window buffer contains sample from both the surface during the transitions, hence, the MLP is not able to distinguish between them. However, this effect disappear after $P$ samples are collected and, as shown in the figure, the estimation rapidly converges towards the ground truth values.

\subsubsection{Closed Loop Results}

\begin{table*}[htb]
	\centering
	\resizebox{0.98\textwidth}{!}{%
		\begin{tabular}{|l|l|l|l|l|l|l|l|l|l|l|l|l|}
			\hline
			\multirow{3}{*}{Asphalt} & \multicolumn{4}{c|}{$RMSE$} & \multicolumn{4}{c|}{Distance (Meters)} & \multicolumn{4}{c|}{Braking time (Seconds)} \\ \cline{2-13} 
			& \multicolumn{2}{c|}{MLP} & \multicolumn{2}{c|}{RLS} & \multicolumn{2}{c|}{MLP} & \multicolumn{2}{c|}{RLS} & \multicolumn{2}{c|}{MLP} & \multicolumn{2}{c|}{RLS} \\ \cline{2-13} 
			& \multicolumn{1}{c|}{PI} & \multicolumn{1}{c|}{SMC} & \multicolumn{1}{c|}{PI} & \multicolumn{1}{c|}{SMC} & \multicolumn{1}{c|}{PI} & \multicolumn{1}{c|}{SMC} & \multicolumn{1}{c|}{PI} & \multicolumn{1}{c|}{SMC} & \multicolumn{1}{c|}{PI} & \multicolumn{1}{c|}{SMC} & \multicolumn{1}{c|}{PI} & \multicolumn{1}{c|}{SMC} \\ \hline
			D & \textbf{0.0070} & \textbf{0.0070} & 0.0120 & \textbf{0.0070} & 280.8347 & \textbf{280.516} & 281.487 & 281.784 & \textbf{6.180} & \textbf{6.180} & 6.190 & 6.195 \\ \hline
			W & \textbf{0.0104} & 0.0105 & 0.0202 & 0.0216 & 406.082 & 405.912 & \textbf{405.793} & 405.939 & 8.980 & \textbf{8.975} & 8.980 & 8.985 \\ \hline
			S & 0.0172 & 0.0172 & \textbf{0.0036} & 0.0799 & 1739.4 & 1740.000 & \textbf{1733.700} & 1771.900 & 38.610 & 38.615 & \textbf{38.495} & 39.355 \\ \hline
			D$\rightarrow$S$\rightarrow$D & 0.0116 & \textbf{0.0102} & 0.1094 & 0.1106 & 382.508 & \textbf{382.166} & 388.825 & 424.357 & \textbf{8.010} & 8.0150 & 8.210 & 9.310 \\ \hline
			S$\rightarrow$D$\rightarrow$S & 0.0172 & \textbf{0.0159} & 0.1992 & 0.0797 & \textbf{640.039} & 644.486 & 1337.900 & 1526.100 & \textbf{18.350} & 18.475 & 32.760 & 35.925 \\ \hline
			W$\rightarrow$D$\rightarrow$W & \textbf{0.0069} & 0.0092 & 0.0653 & 0.0560 & 353.358 & \textbf{352.302} & 364.263 & 364.847 & 7.655 & \textbf{7.635} & 8.010 & 8.000 \\ \hline
			D$\rightarrow$W$\rightarrow$D & \textbf{0.0082} & 0.0089 & 0.0653 & 0.0391 & 313.771 & \textbf{312.375} & 317.432 & 317.248 & 7.155 & \textbf{7.135} & 7.280 & 7.255 \\ \hline
			W$\rightarrow$S$\rightarrow$W & 0.0116 & \textbf{0.0125} & 0.2418 & 0.0800 & 518.687 & \textbf{516.136} & 530.236 & 556.100 & 11.420 & \textbf{11.375} & 11.950 & 12.895 \\ \hline
			S$\rightarrow$W$\rightarrow$S & 0.0177 & \textbf{0.0165} & 0.1756 & 0.0552 & 1004.900 & \textbf{1001.400} & 1417.600 & 1502.300 & 25.820 & \textbf{25.760} & 33.770 & 35.430 \\ \hline
			S$\rightarrow$W$\rightarrow$D & 0.0121 & \textbf{0.0120} & 0.1697 & 0.1436 & 538.298 & \textbf{537.571} & 636.579 & 892.028 & 9.690 & \textbf{9.680} & 11.175 & 18.355 \\ \hline
			W$\rightarrow$D$\rightarrow$S & 0.0172 & \textbf{0.0161} & 0.1046 & 0.1152 & 593.928 & \textbf{591.514} & 625.628 & 629.436 & 18.375 & \textbf{18.330} & 19.320 & 19.410 \\ \hline
			D$\rightarrow$S$\rightarrow$W & 0.0123 & \textbf{0.0122} & 0.0696 & 0.0922 & 427.638 & \textbf{423.989} & 428.758 & 455.905 & 10.065 & \textbf{10.020} & 10.090 & 11.180 \\ \hline
			S$\rightarrow$D$\rightarrow$W & 0.0133 & \textbf{0.0130} & 0.1977 & 0.3248 & \textbf{575.939} & 576.856 & 774.532 & 807.920 & \textbf{10.890} & \textbf{10.890} & 14.295 & 15.145 \\ \hline
			W$\rightarrow$S$\rightarrow$D & \textbf{0.0121} & 0.0122 & 0.5228 & 0.1124 & 484.147 & \textbf{483.461} & 516.051 & 564.871 & \textbf{10.100} & 10.110 & 11.425 & 13.220 \\ \hline
			D$\rightarrow$W$\rightarrow$S & 0.0170 & \textbf{0.0157} & 0.128 & 0.1291 & 500.693 & \textbf{496.981} & 514.565 & 513.717 & 16.425 & \textbf{16.335} & 16.970 & 16.910 \\ \hline
		\end{tabular}%
	}
	\caption{Performance of MLP under closed loop control schemes.  \vspace{-0.5cm}}
	\label{tab:Closedloop}
\end{table*}

Finally, Table \ref{tab:Closedloop} reports the results obtained in the closed loop tests. The estimation performance (RMSEs in the leftmost four columns of the data) shows a trend similar to the open loop tests: the MLP outperforms the RLS in most of the cases. Hence, by using the proposed neural network estimator it is possible to considerably improve the efficiency of the braking procedure, as proven by the comparison on the distance traveled and the braking time. This is well demonstrated, for example, in the S$\rightarrow$W$\rightarrow$S and the S$\rightarrow$D$\rightarrow$S tests, where the controller based on the RLS approach requires up to $700$ meters and $16$ seconds more than the MLP to stop the aircraft. 

Furthermore, by comparing the performance obtained with the PI and the SMC controllers, it can be observed that, in general, their performance are comparable. Thus, this proves that the proposed MLP estimation strategy can be easily coupled with different control strategies.

A qualitative comparison between the time behavior of the MLP and RLS estimates is provided in Figure \ref{fig:closed_loop}. Similarly to the open loop case, the MLP achieves lower errors and the road transitions do not compromise the performance (except from the spike during the road switch, as explained in the previous section). Conversely, the RLS is heavily affected by surface transitions and, in most cases, the estimate diverges and becomes unreliable.




\section{Conclusions} \label{sec:conclusions}

This work proposes a novel data-driven strategy to estimate the optimal slip value to perform efficient braking control. The approach exploits a neural network architecture that can detect the road conditions by processing sequences of slip-friction pairs. The experiments are performed by simulating the landing of an aircraft over un unknown surface whose road conditions change during the braking operation. The results clearly show that the proposed MLP-based estimator achieves better performance when compared to a state-of-the-art RLS approach. 
Future work will analyze other types of neural networks. In particular, Recurrent Neural Network (RNN) will be considered to model temporal correlations.


\bibliographystyle{IEEEtran}
\bibliography{braking_references}





\end{document}